\pdfoutput=1

\documentclass[11pt]{article}

\usepackage[]{acl}

\usepackage{times}
\usepackage{latexsym}
\usepackage{graphicx}
\usepackage{comment}
\usepackage{booktabs}
\usepackage{siunitx}
\usepackage{caption}
\usepackage{listings}
\usepackage{color}
\usepackage{array}
\usepackage{longtable}

\usepackage{booktabs}
\usepackage{listings}
\usepackage{color}

\definecolor{codegreen}{rgb}{0,0.6,0}
\definecolor{codegray}{rgb}{0.5,0.5,0.5}
\definecolor{codepurple}{rgb}{0.58,0,0.82}
\definecolor{backcolour}{rgb}{0.95,0.95,0.92}

\lstdefinestyle{mystyle}{
    backgroundcolor=\color{backcolour},   
    commentstyle=\color{codegreen},
    keywordstyle=\color{magenta},
    numberstyle=\tiny\color{codegray},
    stringstyle=\color{codepurple},
    basicstyle=\ttfamily\footnotesize,
    breakatwhitespace=false,         
    breaklines=true,                 
    captionpos=b,                    
    keepspaces=true,                 
    numbers=left,                    
    numbersep=5pt,                  
    showspaces=false,                
    showstringspaces=false,
    showtabs=false,                  
    tabsize=2
}

\NewDocumentCommand{\codeword}{v}{%
\texttt{{#1}}%
}
\newcommand{\dataset}[1]{{\fontfamily{cmss}\selectfont{#1}}}
\usepackage[T1]{fontenc}

\usepackage[utf8]{inputenc}

\usepackage{microtype}

\usepackage{inconsolata}

\title{CodeInsight: A Curated Dataset of Practical Coding Solutions from Stack Overflow}

\author{Nathanaël Beau$^{1,2}$ \and Benoît Crabbé$^1$ \\
       $^1$ Université de Paris, LLF, CNRS, 75013 Paris, France \\ $^2$ onepoint, 29 rue des Sablons, F-75116 Paris, France \\
       \texttt{nathanael.beau.gs@gmail.com} \\
       \texttt{benoit.crabbe@u-paris.fr} }
\begin{document}
\maketitle
\begin{abstract}
We introduce a novel dataset tailored for code generation, aimed at aiding developers in common tasks. Our dataset provides examples that include a clarified intent, code snippets associated, and an average of three related unit tests. It encompasses a range of libraries such as \texttt{Pandas}, \texttt{Numpy}, and \texttt{Regex}, along with more than 70 standard libraries in Python code derived from Stack Overflow. Comprising 3,409 crafted examples by Python experts, our dataset is designed for both model finetuning and standalone evaluation. To complete unit tests evaluation, we categorize examples in order to get more fine grained analysis, enhancing the understanding of models' strengths and weaknesses in specific coding tasks. The examples have been refined to reduce data contamination, a process confirmed by the performance of three leading models: Mistral 7B, CodeLLaMa 13B, and Starcoder 15B. We further investigate data-contamination testing GPT-4 performance on a part of our dataset. The benchmark can be accessed at \url{https://github.com/NathanaelBeau/CodeInsight}.
\end{abstract}

\section{Introduction}

In the dynamic landscape of software engineering, developers frequently confront the challenge of translating conceptual ideas into functional code. While navigating this process, the gap between intention and implementation can often be a hurdle, even for experienced programmers. Traditionally, developers have turned to online resources like Stack Overflow, searching for solutions in natural language to address their specific coding dilemmas.

The emergence of large language models (LLMs) trained on code has heralded a new era in this domain. Innovations like Codex \cite{codex} have revolutionized the field by providing real-time code suggestions in Integrated Development Environments (IDEs). Similarly, models such as ChatGPT and CodeLLAMA \cite{codellama} demonstrate the potential for integrating into IDEs, offering developers context-aware assistance in initiating and refining code, thereby enhancing the efficiency of the software development cycle.

However, the ascent of code generation through LLMs underscores the heightened need for datasets that emphasize precision, context-awareness, and syntactic accuracy. While existing datasets have propelled advancements in this arena, they have limitations. The shift towards LLM-focused datasets has led to a decreased emphasis on traditional training sets, directing attention towards evaluation sets. This shift challenges the training of models from scratch or for specific task fine-tuning. Moreover, while datasets like HumanEval \cite{humaneval} or APPS \cite{apps} provide valuable insights, they often fall short of mirroring the real-world coding challenges developers encounter.

Addressing these gaps, this paper introduces the CodeInsight dataset, a resource specifically tailored for Python code generation. This focus is anchored in Python's widespread adoption in key sectors like data science, machine learning, and web development. The dataset, comprising 3,409 unique, expert-curated Python examples, spans basic programming to complex data science challenges, complete with unit tests for evaluation. The CodeInsight dataset stands out in its ability to provide a nuanced balance between breadth and depth, offering a finely-tuned resource for training and evaluating LLMs in Python code generation. By bridging the gap between natural language and code, CodeInsight presents a tool for understanding and enhancing the capabilities of LLMs in real-world programming contexts.

The dataset provides three primary innovations, uniquely combined within this resource:
\begin{itemize}
    \item It includes a unit test based evaluation, offering a more robust evaluation metric than traditional methods such as BLEU score.
    \item Examples are annotated to facilitate a deeper analysis of its strengths and weaknesses.
    \item It provides a training set in addition to a test set, with each example being manually curated to ensure high quality, supporting efficient fine-tuning.
\end{itemize}

Organized as follows, this paper first details the dataset construction process in Section \ref{sec:dataset-construction}, including our sources, selection criteria, and annotation methods. Section \ref{sec:dataset-stats} presents astatistical analysis of the dataset, highlighting its diverse applications. In Section \ref{sec:baselines}, the dataset's is tested through evaluations using LLM baselines. Lastly, Section \ref{sec:related-works} situates CodeInsight within the broader landscape of code generation datasets.

\section{Dataset Construction}
\label{sec:dataset-construction}

\begin{figure*}
  \includegraphics[width=\linewidth, scale=1.1]{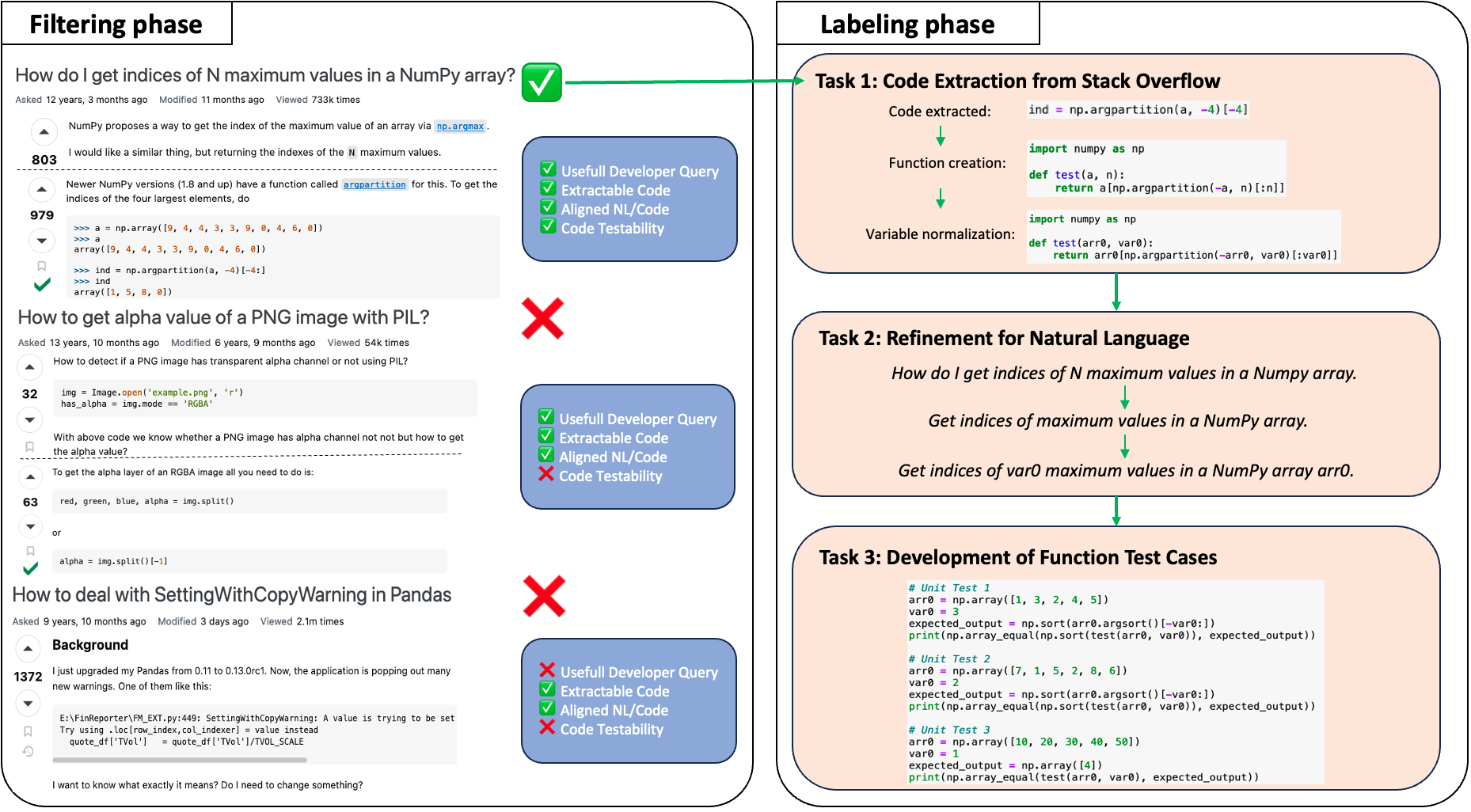}
  \caption{Curation Workflow from Stack Overflow to Dataset - The filtering phase (left) screens questions based on usefulness, code extractability, alignment, and testability, with one example advancing. The labeling phase (right) details the annotation of this example: extracting and standardizing code, refining the question for clarity with normalized terms, and developing unit tests to validate the function.}
  \label{fig:process}
\end{figure*}

Our pipeline for building CodeInsight consists of three pivotal steps. Initially, we identified the sources to retrieve examples. Subsequently, from these sources, we filtered the most relevant natural language-code pairs. The final phase involved annotating these pairs and crafting associated unit tests. This section provides a breakdown of each of these stages. 

\subsection{Data Sources}

To develop a dataset for code generation aimed at aiding development, we prioritize sources that closely mirror real-world development challenges, ensuring a match between natural language and code. We chose Stack Overflow due to its extensive collection of real-world programming questions and solutions, featuring balanced complexity and contributed by a broad and experienced community.

Despite Stack Overflow's extensive collection of developer queries, only 36\% of Python-tagged questions fit the {\it how-to} format essential for our dataset, as identified in \citet{conala}. A {\it 'how-to'} question typically presents a clear, task-oriented query where the developer seeks a method to accomplish a specific programming task. 

To address the challenge of identifying relevant examples, we utilized the CoNaLa dataset \cite{conala}, a curated collection of Python {\it 'how-to'} examples from Stack Overflow. This dataset features 2,379 examples that have been manually reviewed and corrected by annotators, alongside approximately 600,000 unrefined examples ranked by their likelihood of fitting the 'how-to' criteria. Our selection encompassed the 2,379 hand-written examples and the top-ranked 3,121 unrefined examples.

To broaden the scope and applicability of our dataset, we have incorporated an additional 600 samples from Stack Overflow, emphasizing the use of packages like \texttt{Pandas}, \texttt{Numpy}, and \texttt{Regex}. The integration of these packages is a decision to align the dataset with the emergent code generation demands in data science, both in academic research and industry applications. Moreover, \texttt{Regex}’s inclusion enhances the dataset's to accommodate a wider range of specialized problems.

The sourcing procedure began with the elimination of redundancies and the filtration of issues based on a baseline of community engagement—measured by votes and views—and the presence of accepted answers. We then prioritized the problems using a weighted ranking system that accounts for the temporal dimension, recognizing that older issues may naturally garner more attention over time.

Finally, from our selection process, we gathered a total of 7,300 raw examples to serve as the foundation for our dataset.

\subsection{Data Filtering}

 The transition into the data filtering phase necessitates a strategy to select examples from the source, acknowledging that not all contributions from the Stack Overflow community are directly amenable to our goals, as underscored by \citet{conala, ds-1000}. To illustrate, the most upvoted question on pandas is \textit{'How to iterate over rows in a DataFrame in Pandas'}, yet the consensus answer advises against iteration, highlighting the complexity inherent in the selection process.

To navigate these intricacies, we established criteria for inclusion:

\paragraph{Authenticity of Developer Inquiries} Only those questions that present realistic programming scenarios are considered, ensuring the dataset’s relevance to the actual needs of developers.

\paragraph{Direct Extractability of Code} We require that the code snippet can be unambiguously identified and extracted from the accompanying explanatory text. 

\paragraph{Natural Language and Code Alignment}  A robust correspondence between the problem statement and the code solution is necessary for maintaining semantic integrity.

\paragraph{Executable Code Samples}  The code must be functionally valid, capable of running in a designated environment, which is essential for both verifying its effectiveness and constructing unit tests. We decide to exclude code where we need to open or save a file.  \vspace{2mm}

After implementing our filtering process, we refined our initial collection of 7,300 examples to 2,707 distinct problems, constituting about 37\% of the original raw examples. This significant reduction is represented across various sources: from Stack Overflow's \dataset{CoNaLa} dataset, we retained 1,993 out of 5,500 examples; in the \texttt{Pandas}, \texttt{Numpy}, and \texttt{Regex} categories, the numbers were pruned down to 294, 242, and 178 from their respective totals of 600.

The low retention rate in our dataset can be attributed to various factors. For instance, some CoNaLa dataset examples were either non-testable or too specialized, necessitating extensive modification for practical use. Additionally, certain examples offered best practice advice or warnings rather than direct code solutions. The complexity of queries involving advanced libraries like \texttt{Pandas}, \texttt{Numpy}, and \texttt{Regex} also posed challenges. While these queries provide valuable specialized advice on Stack Overflow, they often require significant adaptation for generalization. The second and third examples on the left of Figure \ref{fig:process} illustrate these challenges: one involves extracting features from an image using a library, which is relevant but difficult to test due to the need for incorporating and processing images. The other example from the Pandas library focuses on best practices rather than direct coding solutions, not aligning with the dataset's aim for concrete developer tasks. More examples of what we consider as real-world problems, overly specialized queries, or edge cases are detailed in Appendix \ref{app:filtering_examples}.

\subsection{Data Annotation}
\label{subsec:datalabeling}

Our data annotation workflow is designed to prevent model memorization and instead cultivate problem-solving skills within the generated dataset. Through a multi-stage annotation process presented on the right of Figure \ref{fig:process}, we curate selected examples from the filtering phase into  delineated instances, which are then tested against specially crafted unit tests to ensure their correctness. By refining natural language focusing on the semantic relationships between functions and their descriptions, we diminish the likelihood of models trained on massive datasets to merely replicate solutions seen in their training data, a concern highlighted by \citet{ds-1000} regarding examples sourced from Stack Overflow. To maintain focus and efficiency, annotators are allocated a strict twenty-minute window per example to ensure timely progression and a broad coverage of examples. The ensuing steps show our annotation strategy:

\paragraph{Task 1 - Code Extraction from Stack Overflow} This initial phase entailed the extraction of code solutions from Stack Overflow in response to developers' inquiries. When the question admits more than one valid response, annotators are expected to capture alternate solutions as well, creating a supplementary example for the same intent. Upon extraction, they transform these snippets into a standardized Python function named \texttt{test}, systematically renaming arguments (e.g., \texttt{vari} for variables, \texttt{arri} for arrays, etc. See Appendix \ref{app:normalized-variable-names} for all normalized names). This normalization approach aligns with \citet{conala}, recognizing, as pointed out by \citet{bertranx}, the significant influence this method has on models performance.

\paragraph{Task 2 - Refinement for Natural Language and Code Consistency} During this stage, annotators refined the natural language descriptions to precisely correspond with the \texttt{test} function created in Task 1. The challenge lay in harmonizing the language descriptions with the Python code's logic, ensuring they are concise yet informative. Annotators were also tasked with incorporating normalized argument names into these descriptions to bolster the dataset’s internal coherence and force the alignment.

\paragraph{Task 3 - Development of Function Test Cases} The concluding annotation task involved the generation of 3 unique test cases for each \texttt{test} function, designed to rigorously assess the function’s operational integrity and accuracy. These test cases include a normal scenario, an edge case, and an error situation, providing comprehensive coverage. This ensures a thorough yet time-efficient evaluation. Once the test cases have been passed, annotator can proceed the next example. \vspace{2mm}

A team of five data science professionals, each with a minimum of five years of experience, contributed to the labeling of the filtered examples. They managed to complete the annotation in an average time of twelve minutes per example, amounting to a collective annotation effort of over 540 hours.

This process yielded a compendium of 3,409 examples derived from 2,702 distinct problem statements formulated by seasoned developers.

\begin{table*}[htbp]
\centering
\resizebox{\textwidth}{!}{

\begin{tabular}{l|cc|cc|cc|cc}
\toprule
& \multicolumn{2}{c|}{\textbf{Item Count}} & \multicolumn{2}{c|}{\textbf{Avg. Prob Words}} & \multicolumn{2}{c|}{\textbf{Avg. Code Lines}} & \multicolumn{2}{c}{\textbf{Avg. Depth AST}} \\
\textbf{Package} & \textbf{CodeInsight} & \textbf{DS-1000} & \textbf{CodeInsight} & \textbf{DS-1000} & \textbf{CodeInsight} & \textbf{DS-1000} & \textbf{CodeInsight} & \textbf{DS-1000} \\
\midrule
Full dataset & 3,409 & 1,000 & 12.6 $\pm$ 4.3 & 140.0 & 4.6 $\pm$ 2.3 & 3.6 & 8.6 $\pm$ 1.5 & 8.5 \\
\midrule
\texttt{Pandas} & 819 & 291 & 14.1 $\pm$ 4.2 & 184.8 & 3.59 $\pm$ 1.9 & 5.4 & 8.7 $\pm$ 1.4 & 10.7 \\
\texttt{Numpy} & 591 & 220 & 12.2 $\pm$ 3.3 & 137.5 & 5.3 $\pm$ 1.2 & 2.5 & 8.0 $\pm$ 1.4 & 8.1 \\
\texttt{Scikit-learn} & 19 & 115 & 13.8 $\pm$ 5.5 & 147.3 & 8.1 $\pm$ 7.4 & 3.3 & 7.6 $\pm$ 0.7 & 7.6 \\
\texttt{Scipy} & 8 & 106 & 13.0 $\pm$ 4.4 & 192.4 & 5.5 $\pm$ 1.3 & 3.1 & 6.5 $\pm$ 1.8 & 8.3 \\ \midrule
\texttt{NoImport} & 415 & - & 12.1 $\pm$ 4.0 & - & 3.6 $\pm$ 1.9 & - & 8.2 $\pm$ 1.2 & - \\
\texttt{Re} & 241 & - & 12.2 $\pm$ 2.1 & - & 5.5 $\pm$ 0.8 & - & 8.1 $\pm$ 1.4 & - \\
Other & 1309 & - & 12.5 $\pm$ 3.2 & - & 6.1 $\pm$ 2.8 & - & 8.7 $\pm$ 0.9 & - \\
\midrule
\texttt{Matplotlib} & - & 155 & - & 21.1 & - & 3.0 & - & 6.5 \\
\texttt{TensorFlow} & - & 45 & - & 192.4 & - & 3.1 & - & 7.8 \\
\texttt{Pytorch} & - & 68 & - & 133.4 & - & 2.1 & - & 8.2 \\
\bottomrule
\end{tabular}
}
\caption{Comparative Analysis of Package Statistics in CodeInsight and DS-1000 Datasets. Avg. Prob Words consist into the average length of natural language statement. Standard deviations are reported where applicable. "-" indicates the package is not included in the dataset. Other contains 78 distinct packages like \texttt{Itertools}, \texttt{Collections}, \texttt{Operator}, etc.  Detailed statistical data can be found in Appendix \ref{app:stats-dataset}.}
\label{tab:combined-packages-stats}
\end{table*}

\section{Dataset Statistics}
\label{sec:dataset-stats}

This section outlines the statistical framework of our dataset, highlighting the diversity of programming tasks and the complexity of the included code samples. We approach the analysis from two angles: the representation of code libraries and different labels representing the characteristics of code. Key metrics such as item count, average words per natural language problem, and lines per code sample, alongside the depth of Abstract Syntax Tree (AST) to complete code analysis, are presented to give keys of the difficulty of the dataset.

\subsection{Packages Statistics}

The Table \ref{tab:combined-packages-stats} illustrates the scope of CodeInsight, which encompasses a variety of packages. 

A key aspect of CodeInsight is its focus on concise and precise problem descriptions, a departure from datasets that retain extensive problem contexts. This approach is aimed at reducing the word count in problem descriptions without sacrificing clarity and specificity, a crucial factor for effective code generation.

Code complexity is evaluated using two primary quantitative metrics: the mean line count of code alongside the depth of ASTs. The latter, a measure of syntactic structure complexity, serves to augment the insights gained from line count data. Analysis of AST depth within our dataset reveals a trend: more intricate coding structures, characterized by nested or conditional logic, are associated with deeper ASTs, whereas simpler, linear code correlates with shallower ASTs. Notably, across different packages in our dataset, the AST depth remains relatively consistent, with minor variations observed in packages like \texttt{Scipy} and \texttt{Scikit-learn}, potentially attributable to their smaller sample sizes.

Our dataset is compared with the DS-1000 dataset \cite{ds-1000}, the latter comprising 1,000 evaluation instances and utilizing specialized data science tools from Stack Overflow, complete with extensive problem descriptions. Despite similarities in code line counts and complexity, differences in AST depths, particularly in Pandas, indicate nuanced syntactic complexity variances. CodeInsight features a higher average number of unit tests per example, suggesting a more thorough evaluation methodology. Unlike DS-1000, which lacks detailed analysis of code model failures in unit tests, we provide a statistical breakdown of code categories to enhance understanding of model performance in the next Section.


\begin{figure}[ht!]
  \includegraphics[width=\linewidth, scale=0.9]{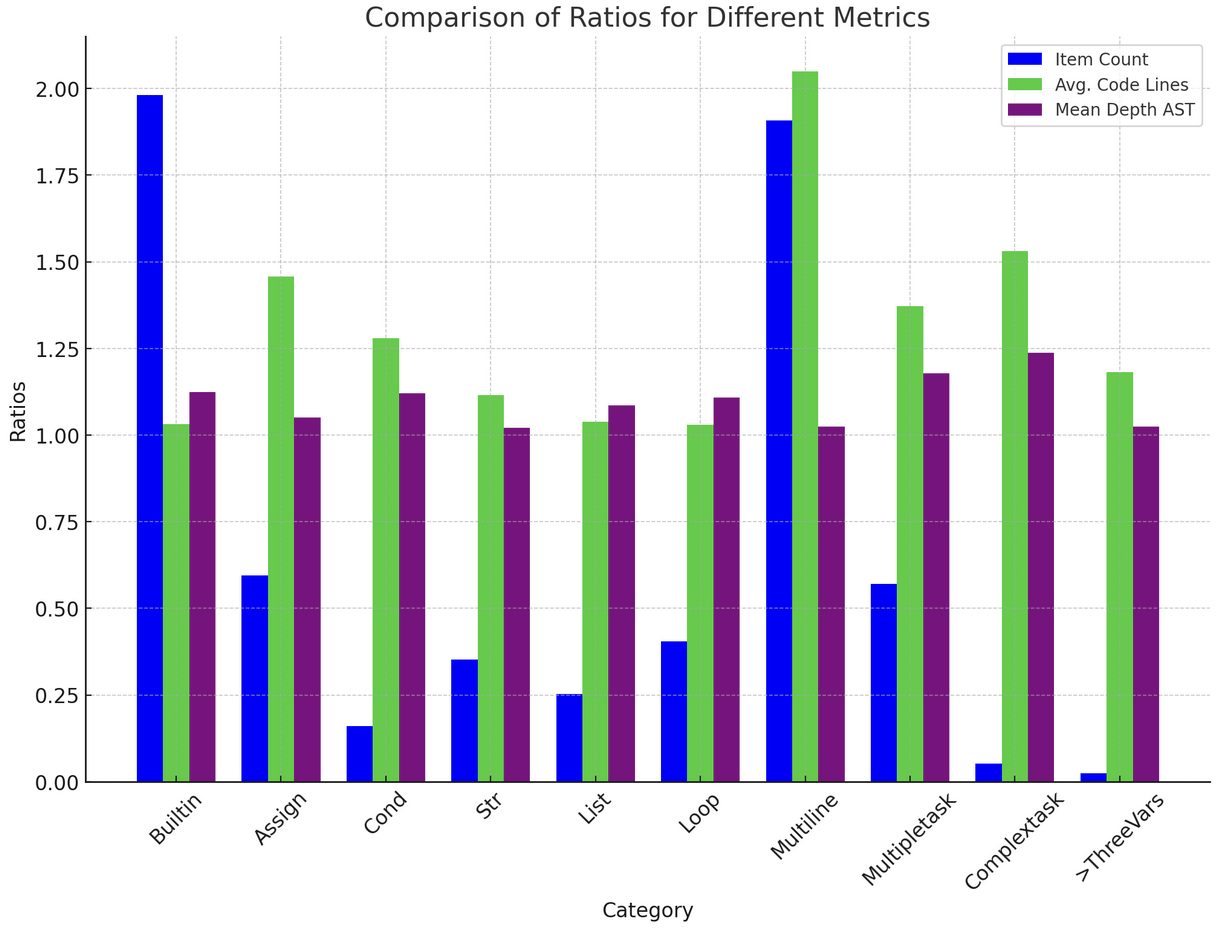}
  \caption{Ratio of positive (belonging to a specific category) to negative (not belonging to the category) examples for each of the 10 distinct {\it Categories} focusing on item count, average code lines and AST depths. Detailed statistical data supporting this analysis can be found in Appendix \ref{app:stats-dataset}.}
  \label{fig:ratios}
\end{figure}

\subsection{Labels Statistics}

In our study, we identified 10 {\it Categories} to enhance our analysis and gain a better understanding of our dataset. These predefined categories provide insights into the conditions under which models were successful or not. These categories vary from basic indicators like \textsc{Builtin} denoting the use of Python's built-in functions, to \textsc{Assign} marking variable assignments. More complex categories include \textsc{Complextask} for codes with multiple imports, and \textsc{>ThreeVars} for functions with over three arguments. Each example in the dataset is binary annotated—marked as positive if it falls under a category and negative otherwise. For a precise definition of all {\it Categories }, refer to Appendix \ref{app:codeinsight_categories}.

Figure \ref{fig:ratios} illustrates the ratio of positive to negative examples for each category to highlight the impact of each category. For example, we compare the \textsc{Assign} category against all examples that do not include variable assignments. Our analysis primarily focuses on the most striking ratios, namely the item count, average code lines and average AST depth, as we found that the unit tests and average problem words exhibit minimal variation across the dataset. Detailed statistical data is provided in Appendix \ref{app:stats-dataset}.

The blue bars in the chart, representing item count ratios, significantly highlight the volume and distribution of data in each category. This showcases the prevalence of certain coding practices; for instance, the \textsc{Builtin} category, with nearly twice as many instances as its counterpart, suggests frequent utilization of built-in functions, indicative of a Pythonic approach in our dataset. In contrast, labels like \textsc{Cond} and \textsc{Loop} exhibit more balanced distributions, reflecting a diverse representation of these elements. Notably, categories such as \textsc{Complextask} and \textsc{>ThreeVars} are less represented, aligning with the expectation of their complexity.

In the context of average code lines, represented by green bars in the graph, specific categories such as \textsc{Complextask}, \textsc{Multipletask}, and \textsc{>ThreeVars} exhibit notably higher ratios. This finding suggests a more intricate and voluminous nature of code associated with these tasks. Contrary to initial expectations, the \textsc{Loop} category does not show an increased number of lines. Further investigation indicates that this outcome can be attributed to the frequent utilization of Python list comprehensions in this category, which typically reduces the number of code lines. In terms of AST depth, it remains relatively consistent for \textsc{Multiline} and \textsc{>ThreeVars} categories. This observation implies that longer codes or handling multiple variables do not necessarily correlate with increased syntactic complexity. However, in the cases of \textsc{Multipletask} and \textsc{Complextask}, there is a correlation between the number of code lines and higher syntactic complexity. For other categories, the complexity levels remain to be stable.

Overall, this analysis underscores the diverse nature of coding practices and the value of categorization in understanding code complexity and coding styles in a nuanced manner. This category-based perspective evaluation on code analysis can be used in general to understand better model mistakes and way to improve model development.

\section{Baselines}
\label{sec:baselines}

In this section, we test our dataset using state-of-the-art LLMs for code generation. Considering the volume and nature of our dataset, we explore various model evaluation methodologies. Initially, we employ a zero-shot evaluation framework, augmenting it with pre-prompts to align better with our specialized task. Subsequently, we experiment diverse partitioning strategies of the dataset for model fine-tuning, followed by evaluation on the remaining data. Additionally, we conduct a comparative performance analysis of GPT-4 on a subset of CoNaLa examples and their modified versions by our annotators, to understand the impact of language and code remodeling relatively to data contamination.

\subsection{Experimental Setup}

\paragraph{Models}

We evaluate the following pre-trained language models: Mistral 7B \cite{mistral} ; CodeLLAMA 13B \cite{codellama} and Starcoder 15B \cite{starcoder}. These models have been selected to provide a perspective on the scalability of model performance in relation to their size and the intricacies of code understanding and generation.

\paragraph{Evaluation Metrics}

We follow \citet{ds-1000} and measure the execution accuracy using the pass@1 metric i.e. we generate one code and test it against all unit tests. We also use the BLEU score \cite{bleu} and the codeBLEU score \cite{codebleu} to complete our evaluation. 

\paragraph{Model input}
For evaluation, we give to the model the intent in natural language and its associated function header with its arguments. Once the generation is finished, we automatically detect the end of the function -when it exists- to get the whole code and test it.

\subsection{Prompting Evaluation}

\begin{table}[h]
\centering
\resizebox{0.48\textwidth}{!}{%
\begin{tabular}{lrrr}
\toprule
 & \textbf{Mistral} & \textbf{CodeLLAMA} & \textbf{Starcoder} \\
\midrule
Without Prompt & 4.7\% & 44.7\% & \textbf{45.1\%} \\
First Prompt & 4.9\% & 40.3\% & \textbf{45.1\%} \\
Second Prompt & 10.1\% & \textbf{48.1\%} & 46.8\% \\
\bottomrule
\end{tabular}
}
\caption{Baselines result varying prompt method. We report the percentage of all unit tests passed (pass@1 score).}
\label{tab:prompt-eval}
\end{table}

\paragraph{Without prompt}
Initially, the models were evaluated using the entire dataset without any additional context added to the natural language intent. The results, as presented in the Table \ref{tab:prompt-eval}, indicate a contrast in performance. Mistral showed notably lower efficiency compared to CodeLLAMA and Starcoder, which both passed nearly 45\% of the unit tests. A key observation was the absence of a return statement in a significant proportion of the generated code. While Python allows for scenarios where not returning an explicit value is acceptable, such as actions or modifications without a return value, our dataset did not align with these scenarios. Mistral particularly exhibited a tendency (25\% of the cases) to end functions with print statements instead of return statements, affecting its accuracy.

\paragraph{First prompt}
In an attempt to steer the models towards generating return statements for development aid tasks, a pre-prompt was introduced: {\it “You are a powerful code generation model. Your job is to convert a given natural language prompt into Python function code and return the result.”}. Surprisingly, this prompt only marginally improved Mistral's performance, with a slight increase in return statement generation. However, it did not significantly affect the performance of CodeLLAMA and Starcoder. Notably, CodeLLAMA's performance even dropped to 40\%, indicating that this prompting method might not be optimal.

\paragraph{Second prompt}
Aiming to further encourage the generation of return statements, a different prompt, {\it “Return the Result.”} was added to the end of the natural language intent. This change led to an overall improvement in performance across all models, with CodeLLAMA outperforming Starcoder. Mistral, although still lagging, showed an improvement, successfully passing 10.1\% of the unit tests.

\subsection{Fine-Tuning Evaluation}

This segment delves into fine-tuning configurations to discern their impact on model efficacy.

\paragraph{Splitting Method}

\begin{table}
\centering
\resizebox{0.45\textwidth}{!}{%
\begin{tabular}{c|ccc}
\toprule
Split & Pass@1 & BLEU & codeBLEU \\
\midrule
20-80 &   $48.9 \pm 0.6\%  $  &  $50.0 \pm 0.2  $ & $42.5 \pm 0.1 $  \\
40-60 &  $  52.6 \pm 0.8 \%  $ & $58.1 \pm 0.4 $   & $48.8 \pm 0.4  $ \\
60-40 &  $53.4 \pm 1.0\%  $ & $57.9 \pm 0.8  $   &  $48.8 \pm 0.7  $\\
80-20 &   $53.1 \pm 1.7\%  $  &  $57.9 \pm 1.4  $  & $48.6 \pm 1.2  $ \\
\bottomrule
\end{tabular}
}
\caption{Scores for Different Splits of CodeLLaMA over five different seed. We report the mean and standard deviation for each metric.}
\label{tab:performance_splits}

\end{table}

\begin{table*}[htbp]
\centering
\resizebox{\textwidth}{!}{%
\begin{tabular}{
  l
  S[table-format=4.0]
  l
  S[table-format=2.1]
  S[table-format=3.1]
  S[table-format=2.1]
  l
  l
}
\toprule
\textbf{Dataset} & {\textbf{Problems}} & \textbf{Evaluation} & {\textbf{Avg. Test Cases}} & {\textbf{Avg. P Words}} & {\textbf{Avg. Lines of Code}} & \textbf{Data Source} & \textbf{Train Set} \\
\midrule
HumanEval & 164 & Test Cases & 7.7 & 23.0 & 6.3 & Hand-Written & No \\
MBPP & 974 & Test Cases & 3.0 & 15.7 & 6.7 & Hand-Written & No\\
APPS & 5000 & Test Cases & 13.2 & 293.2 & 18.0 & Competitions & Yes \\
\midrule
JulCe & 1981 & Exact Match + BLEU & {--} & 57.2 & 3.3 & Notebooks & No\\
DSP & 1119 & Test Cases & 2.1 & 71.9 & 4.5 & Notebooks & No\\
CoNaLa & 500 & BLEU & {--} & 13.8 & 1.1 & StackOverflow & Yes\\
Odex & 945 & Test Cases & 1.8 & 14.5 & 3.9 & Stack Overflow & No\\
DS-1000 & 1000 & Test Cases  & 1.6 & 140.0 & 3.6 & StackOverflow & No\\
\midrule
CodeInsight & 1860 & Test Cases & 3.0 & 12.6 & 4.7 & StackOverflow & Yes\\
\bottomrule
\end{tabular}
}
\caption{Comparison of Test Set Statistics for CodeInsight with recent Code Generation Datasets}
\label{table:comparison-dataset-statistics}
\end{table*}

For the assembly of our test subset, we curated a collection of 3,094 unique problems, each along with at least three unit tests to ensure an assessment of model performance. This selection criterion is grounded in the necessity for test case coverage, which is important in evaluating model robustness. Out of this repository, we allocated different subset to evaluate the need of a train set to perform on test set. 

\paragraph{Fine-Tuning Details}

We finetuned using Lora with $r = 16$ and $\alpha = 16$. The LoRA layer incorporated a dropout rate of 0.05 and was configured without bias adjustments. The batch size was established at 128, encompassing a warmup phase of 100 steps and an overall training regimen of 400 steps. The learning rate was set at \( 3 \times 10^{-5} \), with the optimization executed using the AdamW algorithm. To optimize computational efficiency, training was conducted using half-precision computation (FP16) on an a100 GPU with 40GB memory.

We crafted four distinct training/test splits - 20-80, 40-60, 60-40, and 80-20 - to fine-tune the CodeLLaMa model.
Each split was evaluated over five different seeds, and the results are depicted in the following table.

In our analysis, we noticed that the performance scores for CodeLLaMa exhibit minimal variation when the training set ranges between 40\% to 80\%. Interestingly, these scores surpass those achieved through prompting alone. It appears that fine-tuning with just 20\% of the dataset approaches the performance levels seen with prompting methods, yet it falls short by approximately 4 percentage points in the pass@1 metric and at least 6 points in both BLEU and codeBLEU scores. Given our objective to maximize the utilization of unit tests, we have determined that a 40-60 split represents the most optimal division for the final configuration of the CodeInsight dataset. This decision is grounded in achieving a balanced approach between training efficacy and test coverage.

\subsection{Results}

\begin{table}[htbp]
\centering
\resizebox{0.48\textwidth}{!}{%
\begin{tabular}{|l|c|c|c|c|}
\hline
{Category} & {Total} & {Starcoder} & {CodeLLAMA} & {Mistral} \\
\hline
Full Dataset & 1860 & 52.5\% & \textbf{53.1\%} & 38.4\% \\ \hline
\multicolumn{5}{|c|}{\textit{Labels}} \\
\hline
\textsc{MultiLine} & 1258 & \textbf{51.8\%} & 50.2\% & 42.0\% \\
\textsc{Assign} & 703 & 47.0\% & \textbf{48.2\%} & 40.5\% \\
\textsc{MultipleTask} & 692 & \textbf{44.5\%} & 42.2\% & 39.8\% \\
\textsc{Builtin} & 1292 & \textbf{51.2\%} & 49.8\% & 41.9\% \\
\textsc{Cond} & 260 & 46.7\% & \textbf{47.6\%} & 38.3\% \\
\textsc{Loop} & 573 & \textbf{48.9\%} & 47.8\% & 40.4\% \\
\textsc{List} & 408 & 49.0\% & \textbf{49.5\%} & 41.2\% \\
\textsc{>ThreeVars} & 47 & \textbf{53.5\%} & 53.1\% & 42.3\% \\
\textsc{ComplexTask} & 90 & \textbf{35.6\%} & 34.5\% & 23.1\% \\
\hline
\multicolumn{5}{|c|}{\textit{Packages}} \\
\hline
\texttt{Pandas} & 458 & \textbf{56.0\%} & 55.2\% & 44.8\% \\
\texttt{Numpy} & 335 & \textbf{53.6\%} & 52.8\% & 43.2\% \\
\texttt{NoImport} & 775 & \textbf{54.1\%} & 53.9\% & 44.0\% \\
\texttt{Regex} & 133 & 37.5\% & \textbf{38.3\%} & 26.2\% \\
\hline
\end{tabular}
}
\caption{Baselines Result on final Test Set split 40-60. We report the pass@1 for all models.}
\label{tab:test-set-result}
\end{table}
 Finally, we chose the 40-60 split to perform our final evaluation on our baselines. We report the result in Table \ref{tab:test-set-result}. The Table highlights that fine-tuning has a varied impact on different models. Fine-tuning yields comparable outcomes for Starcoder and CodeLLaMa, each passing slightly over half of the problems. Notably, Starcoder excels in complex tasks like \textsc{ComplexTask} and \textsc{>ThreeVars}, though it drops to 30\% in logical complex tasks. Regex, being a distinct language, poses challenges for all models. Interestingly, Mistral shows significant improvement post-finetuning, adapting well to the task with 38.4\% test pass rate. However, Mistral struggles with complex tasks and Regex, likely due to its non-code-specific pre-training, unlike the other two models.

We provide a deeper error analysis of CodeLLaMa in Appendix \ref{app:erroranalysis}.

\subsection{Exploring remodeling relatively to data contamination}

In Section \ref{subsec:datalabeling}, we detail our approach to mitigate data contamination by rephrasing natural language intents and converting code snippets into function formats. Out of 2,379 CoNaLa handwritten examples, we annotated 812 for analysis. Considering the possibility of these examples being included in GPT-4's training set—a model not open for fine-tuning—we evaluated its zero-shot performance on them, achieving a BLEU score of 58.8.

Further analysis was conducted on the 812 examples post-annotation to assess the impact of our modifications. This evaluation resulted in a BLEU reduction to 47.6.
Remarkably, without fine-tuning, GPT-4 passed 64\% of unit tests for these examples, indicating its effectiveness in understanding natural language.

The performance of GPT-4, despite a drop in BLEU score, suggests its coding capabilities rather than full data leakage from its training phase. The contrast in BLEU scores before and after annotation suggests our approach's impact.  Appendix \ref{app:GPT4_output} presents GPT-4’s predictions, illustrating potential memorization.

\section{Related Works}
\label{sec:related-works}

We introduce a comparative analysis, as detailed in Table \ref{table:comparison-dataset-statistics}, to assess our evaluation set against prevailing code generation datasets. This analysis clusters HumanEval \cite{humaneval}, MBPP \cite{mbpp}, and APPS \cite{apps} due to their emphasis on resolving comprehensive programming challenges. Our dataset, however, is distinguished by its focus on development assistance, which is typically characterized by a lower average line count in the provided code examples, reflecting a different use case compared to the aforementioned datasets.
It is important to note that many datasets are designed primarily for the purpose of evaluating LLMs, which aligns with the prevalent trend in the field. However, the CodeInsight dataset sets itself apart by offering an average number of unit tests per example that exceeds those found in datasets oriented towards data science, such as DSP \cite{dsp}, DS-1000 \cite{ds-1000}, and ODEX \cite{odex}. This difference is largely due to the specific requirements of data science code generation tasks, which often necessitate fewer but more complex test cases, dealing with sophisticated input objects like square matrices, classifiers, or dataframes.

Featuring three distinct unit tests per example, a specialized training set, and predefined labels for in-depth performance analysis, the CodeInsight dataset represents an unparalleled resource. It supports fine-tuning and a comprehensive evaluation of code generation models, offering a novel approach to enhance their development and assessment.

\section{Conclusion}
In conclusion, CodeInsight proposes a new framework for testing code generation, specialized in assisting developers. 
It adeptly links natural language and code in 3,409 problems, providing a robust platform for model training and evaluation. The dataset's strength lies in its diversity, expert annotation, and focus on practical coding scenarios, making it a valuable asset in the intersection of computational linguistics and code generation research. Thanks to its categories, it allows a more precise comprehension of code generation model on this task and is completely compatible with other datasets for development aid.

\section*{Limitations}

The CodeInsight dataset, while innovative, presents several limitations. Firstly, its specialized nature in development aid may not fully represent the broader spectrum of coding challenges. Expert annotations, while valuable, could introduce biases and may not capture diverse coding methodologies. Additionally, the dataset's current scope may limit its adaptability to evolving programming languages and practices. Furthermore, its reliance on Python restricts its applicability across different programming environments. These limitations suggest areas for future expansion and improvement to enhance the dataset's comprehensiveness and applicability in diverse coding contexts.

\bibliography{custom}

\begin{thebibliography}{14}
\expandafter\ifx\csname natexlab\endcsname\relax\def\natexlab#1{#1}\fi

\bibitem[{Austin et~al.(2021)Austin, Odena, Nye, Bosma, Michalewski, Dohan, Jiang, Cai, Terry, Le, and Sutton}]{mbpp}
Jacob Austin, Augustus Odena, Maxwell~I. Nye, Maarten Bosma, Henryk Michalewski, David Dohan, Ellen Jiang, Carrie~J. Cai, Michael Terry, Quoc~V. Le, and Charles Sutton. 2021.
\newblock \href {http://arxiv.org/abs/2108.07732} {Program synthesis with large language models}.
\newblock \emph{CoRR}, abs/2108.07732.

\bibitem[{Beau and Crabb{\'{e}}(2022)}]{bertranx}
Nathana{\"{e}}l Beau and Beno{\^{\i}}t Crabb{\'{e}}. 2022.
\newblock \href {https://doi.org/10.18653/V1/2022.FINDINGS-ACL.173} {The impact of lexical and grammatical processing on generating code from natural language}.
\newblock In \emph{Findings of the Association for Computational Linguistics: {ACL} 2022, Dublin, Ireland, May 22-27, 2022}, pages 2204--2214. Association for Computational Linguistics.

\bibitem[{Chandel et~al.(2022)Chandel, Clement, Serrato, and Sundaresan}]{dsp}
Shubham Chandel, Colin~B. Clement, Guillermo Serrato, and Neel Sundaresan. 2022.
\newblock \href {http://arxiv.org/abs/2201.12901} {Training and evaluating a jupyter notebook data science assistant}.
\newblock \emph{CoRR}, abs/2201.12901.

\bibitem[{Chen et~al.(2021{\natexlab{a}})Chen, Tworek, Jun, Yuan, de~Oliveira~Pinto, Kaplan, Edwards, Burda, Joseph, Brockman, Ray, Puri, Krueger, Petrov, Khlaaf, Sastry, Mishkin, Chan, Gray, Ryder, Pavlov, Power, Kaiser, Bavarian, Winter, Tillet, Such, Cummings, Plappert, Chantzis, Barnes, Herbert{-}Voss, Guss, Nichol, Paino, Tezak, Tang, Babuschkin, Balaji, Jain, Saunders, Hesse, Carr, Leike, Achiam, Misra, Morikawa, Radford, Knight, Brundage, Murati, Mayer, Welinder, McGrew, Amodei, McCandlish, Sutskever, and Zaremba}]{codex}
Mark Chen, Jerry Tworek, Heewoo Jun, Qiming Yuan, Henrique~Pond{\'{e}} de~Oliveira~Pinto, Jared Kaplan, Harrison Edwards, Yuri Burda, Nicholas Joseph, Greg Brockman, Alex Ray, Raul Puri, Gretchen Krueger, Michael Petrov, Heidy Khlaaf, Girish Sastry, Pamela Mishkin, Brooke Chan, Scott Gray, Nick Ryder, Mikhail Pavlov, Alethea Power, Lukasz Kaiser, Mohammad Bavarian, Clemens Winter, Philippe Tillet, Felipe~Petroski Such, Dave Cummings, Matthias Plappert, Fotios Chantzis, Elizabeth Barnes, Ariel Herbert{-}Voss, William~Hebgen Guss, Alex Nichol, Alex Paino, Nikolas Tezak, Jie Tang, Igor Babuschkin, Suchir Balaji, Shantanu Jain, William Saunders, Christopher Hesse, Andrew~N. Carr, Jan Leike, Joshua Achiam, Vedant Misra, Evan Morikawa, Alec Radford, Matthew Knight, Miles Brundage, Mira Murati, Katie Mayer, Peter Welinder, Bob McGrew, Dario Amodei, Sam McCandlish, Ilya Sutskever, and Wojciech Zaremba. 2021{\natexlab{a}}.
\newblock \href {http://arxiv.org/abs/2107.03374} {Evaluating large language models trained on code}.
\newblock \emph{CoRR}, abs/2107.03374.

\bibitem[{Chen et~al.(2021{\natexlab{b}})Chen, Tworek, Jun, Yuan, de~Oliveira~Pinto, Kaplan, Edwards, Burda, Joseph, Brockman, Ray, Puri, Krueger, Petrov, Khlaaf, Sastry, Mishkin, Chan, Gray, Ryder, Pavlov, Power, Kaiser, Bavarian, Winter, Tillet, Such, Cummings, Plappert, Chantzis, Barnes, Herbert{-}Voss, Guss, Nichol, Paino, Tezak, Tang, Babuschkin, Balaji, Jain, Saunders, Hesse, Carr, Leike, Achiam, Misra, Morikawa, Radford, Knight, Brundage, Murati, Mayer, Welinder, McGrew, Amodei, McCandlish, Sutskever, and Zaremba}]{humaneval}
Mark Chen, Jerry Tworek, Heewoo Jun, Qiming Yuan, Henrique~Pond{\'{e}} de~Oliveira~Pinto, Jared Kaplan, Harrison Edwards, Yuri Burda, Nicholas Joseph, Greg Brockman, Alex Ray, Raul Puri, Gretchen Krueger, Michael Petrov, Heidy Khlaaf, Girish Sastry, Pamela Mishkin, Brooke Chan, Scott Gray, Nick Ryder, Mikhail Pavlov, Alethea Power, Lukasz Kaiser, Mohammad Bavarian, Clemens Winter, Philippe Tillet, Felipe~Petroski Such, Dave Cummings, Matthias Plappert, Fotios Chantzis, Elizabeth Barnes, Ariel Herbert{-}Voss, William~Hebgen Guss, Alex Nichol, Alex Paino, Nikolas Tezak, Jie Tang, Igor Babuschkin, Suchir Balaji, Shantanu Jain, William Saunders, Christopher Hesse, Andrew~N. Carr, Jan Leike, Joshua Achiam, Vedant Misra, Evan Morikawa, Alec Radford, Matthew Knight, Miles Brundage, Mira Murati, Katie Mayer, Peter Welinder, Bob McGrew, Dario Amodei, Sam McCandlish, Ilya Sutskever, and Wojciech Zaremba. 2021{\natexlab{b}}.
\newblock \href {http://arxiv.org/abs/2107.03374} {Evaluating large language models trained on code}.
\newblock \emph{CoRR}, abs/2107.03374.

\bibitem[{Hendrycks et~al.(2021)Hendrycks, Basart, Kadavath, Mazeika, Arora, Guo, Burns, Puranik, He, Song, and Steinhardt}]{apps}
Dan Hendrycks, Steven Basart, Saurav Kadavath, Mantas Mazeika, Akul Arora, Ethan Guo, Collin Burns, Samir Puranik, Horace He, Dawn Song, and Jacob Steinhardt. 2021.
\newblock \href {https://datasets-benchmarks-proceedings.neurips.cc/paper/2021/hash/c24cd76e1ce41366a4bbe8a49b02a028-Abstract-round2.html} {Measuring coding challenge competence with {APPS}}.
\newblock In \emph{Proceedings of the Neural Information Processing Systems Track on Datasets and Benchmarks 1, NeurIPS Datasets and Benchmarks 2021, December 2021, virtual}.

\bibitem[{Jiang et~al.(2023)Jiang, Sablayrolles, Mensch, Bamford, Chaplot, de~Las~Casas, Bressand, Lengyel, Lample, Saulnier, Lavaud, Lachaux, Stock, Scao, Lavril, Wang, Lacroix, and Sayed}]{mistral}
Albert~Q. Jiang, Alexandre Sablayrolles, Arthur Mensch, Chris Bamford, Devendra~Singh Chaplot, Diego de~Las~Casas, Florian Bressand, Gianna Lengyel, Guillaume Lample, Lucile Saulnier, L{\'{e}}lio~Renard Lavaud, Marie{-}Anne Lachaux, Pierre Stock, Teven~Le Scao, Thibaut Lavril, Thomas Wang, Timoth{\'{e}}e Lacroix, and William~El Sayed. 2023.
\newblock \href {https://doi.org/10.48550/ARXIV.2310.06825} {Mistral 7b}.
\newblock \emph{CoRR}, abs/2310.06825.

\bibitem[{Lai et~al.(2023)Lai, Li, Wang, Zhang, Zhong, Zettlemoyer, Yih, Fried, Wang, and Yu}]{ds-1000}
Yuhang Lai, Chengxi Li, Yiming Wang, Tianyi Zhang, Ruiqi Zhong, Luke Zettlemoyer, Wen{-}Tau Yih, Daniel Fried, Sida~I. Wang, and Tao Yu. 2023.
\newblock \href {https://proceedings.mlr.press/v202/lai23b.html} {{DS-1000:} {A} natural and reliable benchmark for data science code generation}.
\newblock In \emph{International Conference on Machine Learning, {ICML} 2023, 23-29 July 2023, Honolulu, Hawaii, {USA}}, volume 202 of \emph{Proceedings of Machine Learning Research}, pages 18319--18345. {PMLR}.

\bibitem[{Li et~al.(2023)Li, Allal, Zi, Muennighoff, Kocetkov, Mou, Marone, Akiki, Li, Chim, Liu, Zheltonozhskii, Zhuo, Wang, Dehaene, Davaadorj, Lamy{-}Poirier, Monteiro, Shliazhko, Gontier, Meade, Zebaze, Yee, Umapathi, Zhu, Lipkin, Oblokulov, Wang, V, Stillerman, Patel, Abulkhanov, Zocca, Dey, Zhang, Moustafa{-}Fahmy, Bhattacharyya, Yu, Singh, Luccioni, Villegas, Kunakov, Zhdanov, Romero, Lee, Timor, Ding, Schlesinger, Schoelkopf, Ebert, Dao, Mishra, Gu, Robinson, Anderson, Dolan{-}Gavitt, Contractor, Reddy, Fried, Bahdanau, Jernite, Ferrandis, Hughes, Wolf, Guha, von Werra, and de~Vries}]{starcoder}
Raymond Li, Loubna~Ben Allal, Yangtian Zi, Niklas Muennighoff, Denis Kocetkov, Chenghao Mou, Marc Marone, Christopher Akiki, Jia Li, Jenny Chim, Qian Liu, Evgenii Zheltonozhskii, Terry~Yue Zhuo, Thomas Wang, Olivier Dehaene, Mishig Davaadorj, Joel Lamy{-}Poirier, Jo{\~{a}}o Monteiro, Oleh Shliazhko, Nicolas Gontier, Nicholas Meade, Armel Zebaze, Ming{-}Ho Yee, Logesh~Kumar Umapathi, Jian Zhu, Benjamin Lipkin, Muhtasham Oblokulov, Zhiruo Wang, Rudra~Murthy V, Jason Stillerman, Siva~Sankalp Patel, Dmitry Abulkhanov, Marco Zocca, Manan Dey, Zhihan Zhang, Nour Moustafa{-}Fahmy, Urvashi Bhattacharyya, Wenhao Yu, Swayam Singh, Sasha Luccioni, Paulo Villegas, Maxim Kunakov, Fedor Zhdanov, Manuel Romero, Tony Lee, Nadav Timor, Jennifer Ding, Claire Schlesinger, Hailey Schoelkopf, Jan Ebert, Tri Dao, Mayank Mishra, Alex Gu, Jennifer Robinson, Carolyn~Jane Anderson, Brendan Dolan{-}Gavitt, Danish Contractor, Siva Reddy, Daniel Fried, Dzmitry Bahdanau, Yacine Jernite, Carlos~Mu{\~{n}}oz Ferrandis, Sean Hughes, Thomas
  Wolf, Arjun Guha, Leandro von Werra, and Harm de~Vries. 2023.
\newblock \href {https://doi.org/10.48550/ARXIV.2305.06161} {Starcoder: may the source be with you!}
\newblock \emph{CoRR}, abs/2305.06161.

\bibitem[{Papineni et~al.(2002)Papineni, Roukos, Ward, and Zhu}]{bleu}
Kishore Papineni, Salim Roukos, Todd Ward, and Wei{-}Jing Zhu. 2002.
\newblock \href {https://doi.org/10.3115/1073083.1073135} {Bleu: a method for automatic evaluation of machine translation}.
\newblock In \emph{Proceedings of the 40th Annual Meeting of the Association for Computational Linguistics, July 6-12, 2002, Philadelphia, PA, {USA}}, pages 311--318. {ACL}.

\bibitem[{Ren et~al.(2020)Ren, Guo, Lu, Zhou, Liu, Tang, Sundaresan, Zhou, Blanco, and Ma}]{codebleu}
Shuo Ren, Daya Guo, Shuai Lu, Long Zhou, Shujie Liu, Duyu Tang, Neel Sundaresan, Ming Zhou, Ambrosio Blanco, and Shuai Ma. 2020.
\newblock \href {http://arxiv.org/abs/2009.10297} {Codebleu: a method for automatic evaluation of code synthesis}.
\newblock \emph{CoRR}, abs/2009.10297.

\bibitem[{Rozi{\`{e}}re et~al.(2023)Rozi{\`{e}}re, Gehring, Gloeckle, Sootla, Gat, Tan, Adi, Liu, Remez, Rapin, Kozhevnikov, Evtimov, Bitton, Bhatt, Canton{-}Ferrer, Grattafiori, Xiong, D{\'{e}}fossez, Copet, Azhar, Touvron, Martin, Usunier, Scialom, and Synnaeve}]{codellama}
Baptiste Rozi{\`{e}}re, Jonas Gehring, Fabian Gloeckle, Sten Sootla, Itai Gat, Xiaoqing~Ellen Tan, Yossi Adi, Jingyu Liu, Tal Remez, J{\'{e}}r{\'{e}}my Rapin, Artyom Kozhevnikov, Ivan Evtimov, Joanna Bitton, Manish Bhatt, Cristian Canton{-}Ferrer, Aaron Grattafiori, Wenhan Xiong, Alexandre D{\'{e}}fossez, Jade Copet, Faisal Azhar, Hugo Touvron, Louis Martin, Nicolas Usunier, Thomas Scialom, and Gabriel Synnaeve. 2023.
\newblock \href {https://doi.org/10.48550/ARXIV.2308.12950} {Code llama: Open foundation models for code}.
\newblock \emph{CoRR}, abs/2308.12950.

\bibitem[{Wang et~al.(2022)Wang, Zhou, Fried, and Neubig}]{odex}
Zhiruo Wang, Shuyan Zhou, Daniel Fried, and Graham Neubig. 2022.
\newblock \href {https://doi.org/10.48550/ARXIV.2212.10481} {Execution-based evaluation for open-domain code generation}.
\newblock \emph{CoRR}, abs/2212.10481.

\bibitem[{Yin et~al.(2018)Yin, Deng, Chen, Vasilescu, and Neubig}]{conala}
Pengcheng Yin, Bowen Deng, Edgar Chen, Bogdan Vasilescu, and Graham Neubig. 2018.
\newblock \href {https://doi.org/10.1145/3196398.3196408} {Learning to mine aligned code and natural language pairs from stack overflow}.
\newblock In \emph{Proceedings of the 15th International Conference on Mining Software Repositories, {MSR} 2018, Gothenburg, Sweden, May 28-29, 2018}, pages 476--486. {ACM}.

\end{thebibliography}

\appendix

\section{Detailed overview of filtering phase}
\label{app:filtering_examples}

We include two tables that analyze the exploitability of examples from the CoNaLa dataset. The Table \ref{table:probaexploinormal} presents the 10 examples with the highest probability of exploitability, highlighting their votes, titles, and whether they are exploitable. The Table \ref{table:exploitabilityCoNaLa} displays a random selection of 10 examples from the same dataset, also detailing their exploitability probability, votes, and titles. 

\begin{table}[htbp]
\centering
\resizebox{0.5\textwidth}{!}{%
\begin{tabular}{c|c|c|c}
 \hline
P(expl) & Vote & Title & Exploitability   \\
\hline
0.87 & +8 & Sort a nested list by two elements  & Yes \\ 
0.85& +61 & Converting integer to list in python  & Yes  \\ 
0.85 & +37 & Converting byte string in unicode string  & Yes  \\
0.85 & +7 & List of arguments with argparse  & No  \\ 
0.84 & +20 & How to convert a Date string to a DateTime object? & Yes/No  \\ 
0.82 & +64 & Converting html to text with Python & Yes   \\ 
0.81 & +8 & Ordering a list of dictionaries in python & Yes   \\ 
0.81 & +4 & Two Combination Lists from One List & No   \\ 
0.80 & +4 & Creating a list of dictionaries in python & No   \\ 
0.79 & +16  & {get index of character in python list}  & Yes \\ \hline
\end{tabular}%
}
\caption{\label{table:probaexploinormal} Exploitability of the 10th examples with highest P(exploitability) from CoNaLa dataset}
\end{table}

\begin{table}[htbp]
\begin{center}
\resizebox{0.5\textwidth}{!}{%
\begin{tabular}{c|c|c|c}
 \hline
P(expl) & Vote & Title & Exploitability   \\
\hline
0.75 & +11 & {How can I plot hysteresis in matplotlib?}  & No  \\ 
0.67 & +499 & {How can I get list of values from dict?}  & Yes  \\ 
0.71 & +7 & {How do I stack two DataFrames next to each other in Pandas?}  & Yes  \\
0.56 & +4 & List sorting with multiple attributes and mixed order & No   \\ 
0.10 & +7 & {Set x-axis intervals(ticks) for graph of Pandas DataFrame} & No  \\ 
0.26 & +6 & {pandas binning a list based on qcut of another list}  & No   \\ 
0.05 & +1989 & {Determine the type of an object?} & Yes   \\ 
0.03 & +11 & {Saving an animated GIF in Pillow}  & No\\
0.02 & +5 & {Quiver or Barb with a date axis} & No   \\ 
0.018 & +6 & {Can't pretty print json from python} & No   \\ 
0.008 & +31 & {For loop - like Python range function} & No \\ 
\hline
\end{tabular}
}
\caption{Exploitability of 10th random from CoNaLa dataset}
\label{table:exploitabilityCoNaLa}
\end{center}
\end{table} 

We provide a detailed description of an accepted example, a rejected example and a borderline case for passing the filtering phase.

\paragraph{Accepted example}

We detailed the accepted example which is the first one on the left of the Figure \ref{fig:process}. This particular example, a query about finding the largest values in a numpy array, demonstrates a typical developer's question due to unfamiliarity with specific numpy functions. Its solution, involving the \texttt{argpartition} function, is directly responsive to the query and easily testable, making it a perfect fit for our dataset.

\paragraph{Rejected example}

The {\it "List sorting with multiple attributes and mixed order"} question on Stack Overflow from Table \ref{table:exploitabilityCoNaLa} presents an excessive level of specificity for inclusion whereas it has a high P(expl) value. This question delves into sorting a list by different attributes of a particular class, emphasizing the specific class's complexity rather than a broader understanding of sorting functions. The high level of detail in both the problem and its solution complicates the extraction of universally applicable code examples. Therefore, including it may not aptly represent the range of coding tasks and challenges.

\paragraph{Edge example}
An example such as {\it "How to convert a Date string to a datetime object?"} presented in Table \ref{table:probaexploinormal} necessitates a more specific reformulation for a precise coding answer, like "How to compare a date string in ISO format to a datetime object." This demands the annotator's understanding of ISO format data. These are considered edge cases in our dataset and are included based on the annotator's expertise, who are constrained to a maximum of 20 minutes per annotation process.

\section{Normalized variable names}
\label{app:normalized-variable-names}

\begin{center}
\begin{table}[htbp]
\begin{center}
\resizebox{0.2\textwidth}{!}{%
\begin{tabular}{ll}\hline
\textsc{Label} & \textsc{Condition}\\\hline
\texttt{vari} & Variable \\
\texttt{dicti}  & Dictionary  \\
\texttt{arri}   & Array \\
\texttt{dfi} & Dataframe \\
\texttt{stri} & String \\
\texttt{lsti} & List\\
\texttt{mati} & Matrix \\
\texttt{inti} & Int\\

\hline
\end{tabular}
}
\caption{\label{tab:label-examples} List of normalized variable names used in our dataset}
\end{center}
\end{table}
\end{center}

The Table \ref{tab:label-examples} outlines the standardized variable names utilized in the dataset, such as \texttt{vari} for 'Variable' and \texttt{dicti} for 'Dictionary', where i correspond to the number of the element appearing. This approach also allows for evaluating model efficacy with or without these normalized names. Note that \texttt{vari} is employed universally, even when alternatives might be applicable, without affecting test outcomes.

\section{Code Categories}
\label{app:codeinsight_categories}

\begin{center}
\begin{table}[htbp]
\resizebox{0.5\textwidth}{!}{%
\begin{tabular}{ll}\hline
\textbf{Label} & \textbf{Condition Description}\\\hline
\textsc{assign} & Includes variable assignment.\\
\textsc{builtin}  & Uses a built-in function.\\
\textsc{cond} & Has conditional statement(s).\\
\textsc{loop} & Contains `for` or `while` loops.\\
\textsc{str} & Performs string operation(s).\\
\textsc{list} & Uses list method(s).\\ \hline
\textsc{MultiLine} & Code exceeds two lines.\\
\textsc{MultipleTask} & Has $\geq$3 other Labels.\\
\textsc{>ThreeVars} & Function with $>$3 parameters.\\
\textsc{ComplexTask} & Has $\geq$2 imports\\
\hline
\end{tabular}
}
\caption{Detailed Labels for Automated Annotation}
\label{tab:label-examples-detailed}
\end{table}
\end{center}

\section{CodeInsight Statistics}
\label{app:stats-dataset}

The two tables provide a detailed statistical analysis of the CodeInsight dataset, breaking down by Packages and Labels. The Table \ref{tab:packages-stats} covers various Python packages like Pandas, Numpy, and Regex, detailing the item count, average problem words, code lines, and unit tests. The second Table \ref{tab:label-stats} analyzes different labels presented in Appendix \ref{app:codeinsight_categories} such as Builtin, Assign, Cond, and others, also including their item count and average metrics. Both tables gives insight on the dataset's complexity and diversity into the typical problem structure and testing framework associated with different programming constructs and packages.

\begin{table}[htbp!]
\centering
\resizebox{0.5\textwidth}{!}{
\begin{tabular}{lcccc}
\toprule
& \textbf{Item Count} & \textbf{Avg. Prob Words} & \textbf{Avg. Code Lines} & \textbf{Avg. Unit Tests} \\
\midrule
Full dataset & 3,409 & $12.6 \pm 4.3$ & $4.6 \pm 2.3$ & $3.0 \pm 0.4$ \\ \hline
\texttt{NoImport} & 415 & $12.1 \pm 4.0$ & $3.6 \pm 1.9$ & $3.0 \pm 0.4$ \\
\texttt{Pandas} & 819 & $14.1 \pm 4.2$ & $5.4 \pm 1.8$ & $3.0 \pm 0.2$ \\
\texttt{Numpy} & 591 & $12.2 \pm 3.3$ & $5.3 \pm 2.0$ & $3.0 \pm 0.2$ \\
\texttt{Re} & 241 & $12.2 \pm 2.1$ & $5.5 \pm 0.8$ & $3.0 \pm 0.2$ \\
\texttt{Scikit-learn} & 19 & $13.8 \pm 5.5$ & $8.1 \pm 7.4$ & $3.0 \pm 0.0$ \\
\texttt{Scipy} & 8 & $13.0 \pm 4.4$ & $5.5 \pm 1.3$ & $3.0 \pm 0.0$ \\
\texttt{Itertools} & 55 & $11.8 \pm 3.5$ & $6.4 \pm 3.1$ & $3.0 \pm 0.4$ \\
\texttt{Collections} & 39 & $13.1 \pm 3.5$ & $6.8 \pm 2.6$ & $3.0 \pm 0.2$ \\
\texttt{Operator} & 43 & $13.4 \pm 3.0$ & $5.0 \pm 1.4$ & $3.2 \pm 0.5$ \\
\texttt{String} & 8 & $9.0 \pm 1.8$ & $5.8 \pm 1.1$ & $3.0 \pm 0.0$ \\
\texttt{Random} & 14 & $12.0 \pm 2.0$ & $5.4 \pm 2.4$ & $2.9 \pm 0.5$ \\
\texttt{Math} & 8 & $13.1 \pm 4.7$ & $6.0 \pm 1.9$ & $2.9 \pm 0.3$ \\
\bottomrule
\end{tabular}
}
\caption{Statistical analysis of Packages in CodeInsight. We report including Item Count, Average Problem Words, Code Lines, and Unit Tests with Standard Deviations.}
\label{tab:packages-stats}
\end{table}

\begin{table}[htbp!]
\centering
\resizebox{0.5\textwidth}{!}{
\begin{tabular}{lcccc}
\toprule
& \textbf{Item Count} & \textbf{Avg. Prob Words} & \textbf{Avg. Code Lines} & \textbf{Avg. AST depth} \\
\midrule
Full dataset & 3402 & $12.6 \pm 4.3$ & $4.6 \pm 2.3$ & $3.0 \pm 0.4$ \\ \hline
\textsc{Builtin} & 2261 & $12.7 \pm 3.8$ & $4.7 \pm 2.2$ & $8.7 \pm 1.5$ \\
\textsc{NoBuiltin} & 1141 & $12.4 \pm 3.6$ & $4.6 \pm 1.4$ & $7.7 \pm 1.2$ \\
\textsc{Assign} & 1269 & $13.2 \pm 3.9$ & $5.8 \pm 2.4$ & $8.6 \pm 1.4$ \\
\textsc{NoAssign} & 2133 & $12.3 \pm 3.6$ & $4.0 \pm 1.4$ & $8.2 \pm 1.5$ \\
\textsc{Cond} & 471 & $13.4 \pm 3.8$ & $5.8 \pm 2.9$ & $9.2 \pm 1.3$ \\
\textsc{NoCond} & 2931 & $12.5 \pm 3.8$ & $4.5 \pm 1.8$ & $8.2 \pm 1.4$ \\
\textsc{Str} & 885 & $12.8 \pm 3.5$ & $5.1 \pm 2.0$ & $8.5 \pm 1.6$ \\
\textsc{NoStr} & 2517 & $12.6 \pm 3.9$ & $4.5 \pm 2.0$ & $8.3 \pm 1.5$ \\
\textsc{List} & 685 & $12.8 \pm 3.8$ & $4.8 \pm 3.0$ & $8.9 \pm 1.3$ \\
\textsc{NoList} & 2717 & $12.6 \pm 3.8$ & $4.7 \pm 1.6$ & $8.2 \pm 1.5$ \\
\textsc{Loop} & 981 & $12.8 \pm 3.8$ & $4.8 \pm 2.8$ & $9.0 \pm 1.3$ \\
\textsc{NoLoop} & 2421 & $12.5 \pm 3.8$ & $4.6 \pm 1.5$ & $8.2 \pm 1.5$ \\ \hline
\textsc{Multiline} & 2232 & $12.8 \pm 3.7$ & $5.5 \pm 1.9$ & $8.5 \pm 1.5$ \\
\textsc{NoMultiline} & 1170 & $12.2 \pm 3.9$ & $2.7 \pm 0.5$ & $8.3 \pm 1.4$ \\
\textsc{Multipletask} & 1236 & $13.2\pm 3.8$ & $5.6 \pm 2.5$ & $8.9 \pm 1.5$ \\
\textsc{NoMultipleTask} & 2166 & $12.3 \pm 3.7$ & $4.1 \pm 1.4$ & $7.8 \pm 1.4$ \\
\textsc{Complextask} & 169 & $13.2 \pm 3.8$ & $7.0 \pm 2.8$ & $10.4 \pm 2.3$ \\
\textsc{NoComplexTask} & 3233 & $12.6 \pm 3.8$ & $4.6 \pm 1.9$ & $8.3 \pm 1.5$ \\
\textsc{>ThreeVars} & 82 & $16.9 \pm 4.2$ & $5.5 \pm 1.2$ & $8.6 \pm 1.4$ \\
\textsc{<=ThreeVars} & 3320 & $12.5 \pm 3.7$ & $4.7 \pm 2.0$ & $8.4 \pm 1.5$ \\
\bottomrule
\end{tabular}
}
\caption{Statistical analysis of Labels in CodeInsight. We report including Item Count, Average Problem Words, Code Lines, and AST depth with Standard Deviations.}
\label{tab:label-stats}
\end{table}

\section{Code Error Analysis}
\label{app:erroranalysis}

\begin{table*}[htbp]
\noindent
\begin{center}
\begin{tabular}{p{3cm}|p{6cm}|p{6cm}}
\toprule
Statement & Generated Code  & Reference Code  \\
\midrule
{\it Extract decision rules from a scikit-learn decision tree tree0 with feature\_names feature\_names0} &
\begin{lstlisting}
from sklearn.tree import export_text
def test(tree0, feature_names0):
    return export_text(tree0, feature_names0)
\end{lstlisting}

&
\begin{lstlisting}
from sklearn.tree import export_text
def test(tree0, feature_names0):
    return export_text(tree0, feature_names=feature_names0)
\end{lstlisting}
\\
\midrule
{\it Split a string str0 and keep the separators} &
\begin{lstlisting}
import re
def test(str0: str) -> list:
    return re.split(r'\W+', str0)
\end{lstlisting}
&
\begin{lstlisting}
import re
def test(str0: str) -> list:
    return [x for x in re.split(r'(\W+)', str0) if x]
\end{lstlisting}
\\
\midrule
{\it Reset index in a pandas dataframe df0} &
\begin{lstlisting}
import pandas as pd
def test(df0):
    df0.reset_index(inplace=True)
    return df0
\end{lstlisting}
&
\begin{lstlisting}
import pandas as pd
def test(df0):
    return pd.DataFrame(df0.values, columns=df0.columns)
\end{lstlisting}

\\
\bottomrule
\end{tabular}
\caption{Error analysis of CodeLLaMa outputs on our final evaluation set.}
\label{tab:erroranalysis}
\end{center}
\end{table*}

We present in Table \ref{tab:erroranalysis} an error analysis of different CodeLLaMa outputs on our dataset. 

For the first scenario, involving scikit-learn, the task is to generate a textual representation of a decision tree with specified feature names for readability. Despite providing the model with import statements and the function signature, the generated code omits the necessary argument syntax \texttt{feature\_names=feature\_names0}, resulting in a TypeError due to excessive positional arguments. This minor but pivotal error exemplifies a scenario where a developer could struggle with argument specification.

In the second case, the task is to split a string while retaining separators, such as punctuation or special tokens like newline characters. The generated code, however, omits an essential parenthesis in the \texttt{re.split} function, leading to a split that excludes the separators. This highlights the dual complexity of understanding both Python and Regex syntaxes.

The final example presents an annotation discrepancy. It involves resetting the index of a Pandas dataframe without specific instructions on handling the old index. The model correctly employs the \texttt{reset\_index} function, typically retaining the old index as a new column. However, the reference code, and consequently the unit tests, do not preserve the old index. Thus, while the generated code aligns with the stated task, it fails unit tests due to the discrepancy in index handling. This case underscores the need for nuanced dataset analysis and exemplifies the challenges of borderline scenarios in dataset construction.

\section{GPT-4's prediction}
\label{app:GPT4_output}

We investigate data contamination within the CoNaLa and our dataset by examining GPT-4's outputs on both dataset. This can offer information in re-writting example during data annotation phase to mitigate data contamination.
\\

\begin{table}[htbp!]
\begin{tabular}{>{\raggedright}p{0.25\linewidth} >{\raggedright}p{0.32\linewidth} >{\raggedright\arraybackslash}p{0.32\linewidth}}
\toprule
\textbf{Intent} & \textbf{Reference Solution} & \textbf{GPT-4 Prediction} \\
\midrule
{\it Convert a list of integers x into a single integer} & \begin{lstlisting}
r = int(''.join(map(str, x)))
\end{lstlisting} & \begin{lstlisting}
r = int(''.join(map(str, x)))
\end{lstlisting} \\ \midrule
{\it Convert a DateTime string back to a DateTime object of format \%Y-\%m-\%d \%H:\%M:\%S} & \begin{lstlisting}
datetime.strptime('2010-11-13 10:33:54', '%Y-%m-%d %H:%M:%S')
\end{lstlisting} & \begin{lstlisting}
datetime.datetime.strptime(date_string, '%Y-%m-%d %H:%M:%S')
\end{lstlisting} \\ \midrule
{\it Reverse sort dictionary \texttt{d} based on its values} & \begin{lstlisting}
sorted(list(d.items()), key=lambda k_v: k_v[1], reverse=True)
\end{lstlisting} & \begin{lstlisting}
sorted(list(d.items()), key=lambda k_v: k_v[1], reverse=True)
\end{lstlisting} \\
\bottomrule
\end{tabular}\caption{GPT-4's Outputs Comparison for CoNaLa}
\label{tab:gpt4_CoNaLa}
\end{table}

\begin{table}[htbp]
\centering
\begin{tabular}{>{\raggedright}p{0.25\linewidth} >{\raggedright}p{0.32\linewidth} >{\raggedright\arraybackslash}p{0.32\linewidth}}
\toprule
\textbf{Intent} & \textbf{Reference Solution} & \textbf{Prediction} \\
\midrule
{\it Convert a list of integers \texttt{lst0} into a single integer.} & \begin{lstlisting}
def test(lst0):
    return int(''.join(map(str, lst0)))
\end{lstlisting} & \begin{lstlisting}
def test(lst0):
    return int(''.join(map(str, lst0)))
\end{lstlisting} \\ \midrule
{\it Convert a datetime string \texttt{str0} back to datetime object of format \%Y-\%m-\%d \%H:\%M:\%S.} & \begin{lstlisting}
from datetime import datetime

def test(str0):
    return datetime.strptime(str0, "%Y-%m-%dT%H:%M:%S")
\end{lstlisting} & \begin{lstlisting}
from datetime import datetime

def test(str0):
    return datetime.strptime(str0, "%Y-%m-%dT%H:%M:%S")
\end{lstlisting} \\ \midrule
{\it Sort in reversing order the items in dictionary \texttt{dict0} by their first values.} & \begin{lstlisting}
def test(dict0):
    return dict(sorted(dict0.items(), key=lambda item: item[1], reverse=True))
\end{lstlisting} & \begin{lstlisting}
def test(dict0):
    return dict(sorted(dict0.items(), key=itemgetter(1), reverse=True))

\end{lstlisting} \\
\bottomrule
\end{tabular}
\caption{GPT-4's Outputs for CodeInsight}
\label{tab:gpt4_codeinsight}
\end{table}

\paragraph{Discussion}

This discussion presents an analysis of GPT-4's outputs on equivalent examples from the CoNaLa dataset and our dataset, CodeInsight, as detailed in Tables \ref{tab:gpt4_CoNaLa} and \ref{tab:gpt4_codeinsight}.

For the CoNaLa dataset, the analysis of GPT-4's predictions reveals interesting observations for the first and third examples. Specifically, GPT-4 autonomously includes an assignment to variable \texttt{r} in its prediction for the first example, despite the absence of such instruction in the original example. Similarly, in the third example, GPT-4 employs the \texttt{k\_v} variable, an uncommon choice, demonstrating a potential memorization to variable naming.

Conversely, the second example highlights GPT-4's ability to generalize. The model infers the use of \texttt{date\_string} even though the specific datetime format is not explicitly mentioned in the intent, showcasing its adeptness at filling in contextual gaps based on the provided intent.

Regarding the outputs on CodeInsight for the same rewritten intents, GPT-4's accuracy remains consistent for the first example. The second example further underscores GPT-4's precision, where the model's prediction aligns exactly with the reference, attributed to a more clearly defined intent or to a memorization from CoNaLa.

The third example diverges in response using itemgetter instead of lambda function but notably passes our unit tests, illustrating GPT-4's capacity for generating viable alternative solutions. This indicates that despite potential differences, GPT-4's inherent generalization capabilities enable it to offer valid code solutions, reflecting its understanding of programming concepts without memorizing data.

\end{document}